\documentclass{Interspeech2024}

\interspeechcameraready

\title{Textless Dependency Parsing by Labeled Sequence Prediction}

\name[affiliation={1}]{Shunsuke}{Kando}
\name[affiliation={1}]{Yusuke}{Miyao}
\name[affiliation={1}]{Jason}{Naradowsky}
\name[affiliation={1,2}]{Shinnosuke}{Takamichi}

\address{
  $^1$The University of Tokyo, Japan \quad
  $^2$Keio University, Japan}
\email{\{skando,yusuke,narad\}@is.s.u-tokyo.ac.jp, shinnosuke\_takamichi@keio.jp}


\usepackage{url}
\usepackage{float}
\usepackage{booktabs}
\usepackage{multirow}
\usepackage{tikz-dependency}
\usepackage{framed}
\usepackage{amssymb}% http://ctan.org/pkg/amssymb
\usepackage{pifont}% http://ctan.org/pkg/pifont

\newcommand{\cmark}{\ding{51}}%
\newcommand{\xmark}{\ding{55}}%

  {\endMakeFramed}
\begin{document}

\maketitle

% the abstract here must exactly match the abstract entered into the paper submission system

\begin{abstract}
Traditional spoken language processing involves cascading an automatic speech recognition (ASR) system into text processing models.
In contrast, ``textless'' methods process speech representations without ASR systems, enabling the direct use of acoustic speech features.
Although their effectiveness is shown in capturing acoustic features, it is unclear in capturing lexical knowledge.
This paper proposes a textless method for dependency parsing, examining its effectiveness and limitations.
Our proposed method predicts a dependency tree from a speech signal without transcribing, representing the tree as a labeled sequence.
scading method outperforms the textless method in overall parsing accuracy, the latter excels in instances with important acoustic features.
Our findings highlight the importance of fusing word-level representations and sentence-level prosody for enhanced parsing performance.
The code and models are made publicly available\footnote{https://github.com/mynlp/SpeechParser}.
\end{abstract}
\section{Introduction}

Textless NLP\footnote{https://speechbot.github.io/} is an emerging approach for spoken language processing (SLP).
Unlike the conventional method that cascades an ASR system into a text processing model, Textless NLP directly processes speech representations without explicitly transcribing texts.
The textless approach offers advantages by preventing ASR errors from propagating to a downstream model, and by retaining acoustic speech features (such as prosody) which are lost in the transcription process of cascading systems.

Textless NLP has demonstrated its effectiveness in tasks where capturing acoustic features is more important than lexical knowledge, including speech resynthesis~\cite{lakhotiaGenerativeSpokenLanguage2021,polyakSpeechResynthesisDiscrete2021} or emotion conversion~\cite{kreukTextlessSpeechEmotion2022}.
However, it is unclear to what extent a textless method can solve downstream tasks that build upon lexical knowledge (such as word semantics or part-of-speech tag), given its lack of explicit reliance on word-level representations.
This property can be particularly critical in syntactic parsing, where understanding word-level relationships is paramount.
% Furthermore, it is non-trivial to model relationships between words without word-level representations.

\begin{figure}[t]
\centering
\begin{tabular}{c|c}
Wav2tree \cite{pupierEndtoEndDependencyParsing2022} (cascading) & proposed (textless) \\
\hline
\includegraphics[width=.23\textwidth]{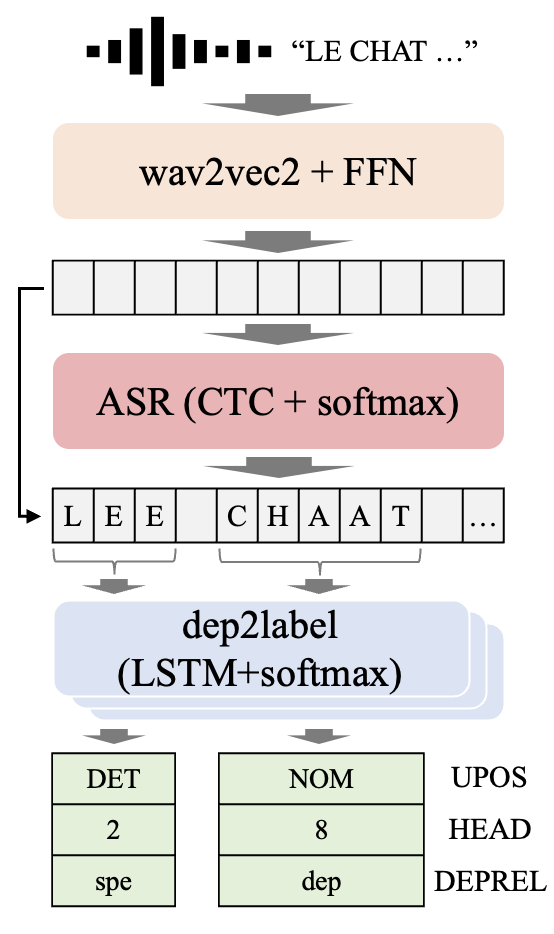} & 
\raisebox{-0.15ex}{\includegraphics[width=.195\textwidth]{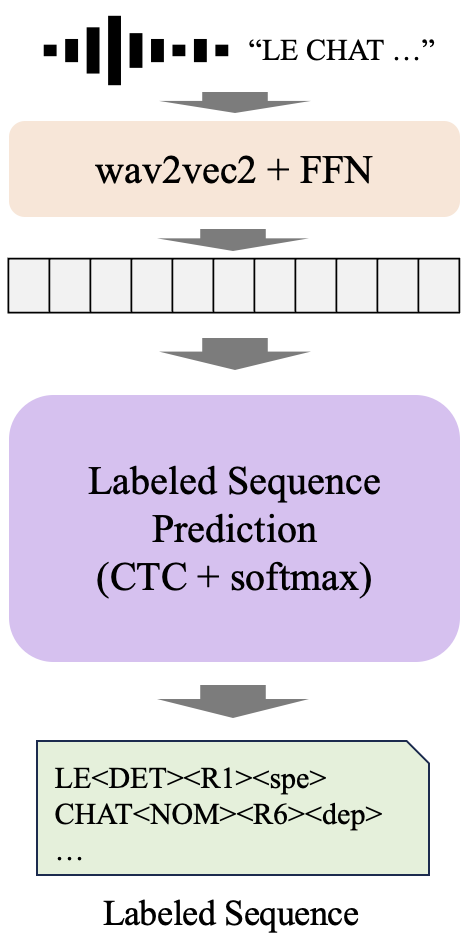}}
\end{tabular}
\caption{
Comparison of Wav2tree~\cite{pupierEndtoEndDependencyParsing2022} (cascading) and the proposed method (textless).
While Wav2tree includes an ASR module, our proposed method directly predicts a dependency tree (represented as a labeled sequence of tokens).
}
\label{fig:overview}
\end{figure}

In this paper, we propose a method for textless dependency parsing and examine its effectiveness and limitations.
Figure~\ref{fig:overview} shows a comparative overview of the cascading and proposed method.
% Unlike the previous cascading method (Wav2tree, \cite{pupierEndtoEndDependencyParsing2022}),
Previous work (Wav2tree, \cite{pupierEndtoEndDependencyParsing2022}) applies the cascading approach for dependency parsing from the speech signal, transcribing the speech, and then utilizing the information of word boundaries for parsing.
In contrast, our proposed method predicts a dependency tree directly from the speech signal, bypassing the step to obtain word-level representations.
The tree is represented as \emph{a labeled sequence}, a concatenation of words and their corresponding annotations (see Figure~\ref{fig:seq} as an example).
% Dependency annotations are represented by special symbols enclosed in angle brackets, such as \texttt{<POS1>}.
This method is inspired by previous work on the sequence-to-sequence model to predict transcription and corresponding linguistic annotations (phonemes and part-of-speech tags) simultaneously~\cite{omachiEndtoendASRJointly2021}.
Dependency parsing is a different task in that dependency relations are not properties of a single word but exist between words (sometimes at long distances).

We empirically compare the cascading and textless methods by evaluating ASR and parsing performance.
In experiments on two languages (French and English), we find that the cascading method outperforms the proposed method overall, particularly in predicting longer dependency relationships.
This suggests that explicitly segmenting a speech at the word boundary is important for enhanced parsing performance.
In contrast, we find that the textless method excels in cases where the important audio feature (such as stress) appears to provide cues for disambiguating the sentence's meaning, such as detecting the main verb of the sentence (i.e. root word).
This suggests that sentence-level prosodic contour may play an important role in parsing.
Our findings suggest the importance of incorporating both word-level representations and sentence-level prosody for improving the parsing performance of speech.

\section{Wav2tree: A Cascading Method}

Previous work proposed Wav2tree~\cite{pupierEndtoEndDependencyParsing2022}, a method for dependency parsing from the speech signal, comprising an ASR model followed by a subsequent parser. 
Wav2tree first extracts the speech representation $\mathbf{X}$ from a signal $S$:
\begin{equation}
    \mathbf{X} = \mathrm{FNN}(f(S)),
\end{equation}
where $\mathbf{X}\in \mathbb{R}^{t\times d}$ (with $t$ denoting the number of frames and $d$ the dimension for the representation), $f$ denotes a feature extractor (pre-trained wav2vec2~\cite{baevskiWav2vecFrameworkSelfSupervised2020}), and FNN denotes a feed-forward neural network.

\subsection{ASR Module}
\label{ssec:asr}

Wav2tree predicts a transcription $\mathbf{w} = w_1 \ w_2 \ \ldots \ w_n$, a sequence of words separated by spaces.
The prediction model $p(\mathbf{w}|\mathbf{X})$ is learned by Connectionist Temporal Classification (CTC) loss \cite{gravesConnectionistTemporalClassification}.

In decoding, given a vocabulary set $\mathcal{V}$, speech representation $\mathbf{X}$ is fed to a linear transformation with softmax, followed by decoders to obtain a transcription:
\begin{align}
    \mathbf{P}_\mathrm{CTC} &= \mathrm{softmax}(\mathbf{X}\mathbf{W}_{\mathrm{CTC}} + b) \\
    \{v_i\}_{i=1}^{n'} &= \mathrm{Dec}_\mathrm{ctc}(\mathbf{P}_\mathrm{CTC}) \ (v_i \in \mathcal{V})\\
    \{w_i\}_{i=1}^{n} &= \mathrm{Dec}_{\mathrm{spm}}(v_i)
\end{align}
where $\mathbf{W}_{\mathrm{CTC}} \in \mathbb{R}^{t\times|\mathcal{V}|}$ is a weight matrix for CTC and $b\in\mathbb{R}^{|\mathcal{V}|}$ is a bias term.
The probability matrix $\mathbf{P}_\mathrm{CTC}$ is first decoded by the CTC decoder ($\mathrm{Dec}_\mathrm{ctc}$), and subsequently by the SentencePiece decoder~\cite{kudoSentencePieceSimpleLanguage2018} ($\mathrm{Dec}_\mathrm{spm}$).
Note that the length of the token sequence decoded by $\mathrm{Dec}_\mathrm{ctc}$ (i.e., $n'$) is not equal to $n$ in general, as each word is decoded by combining tokens from the SentencePiece vocabulary.

\begin{figure}
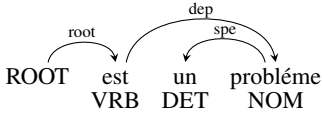

\centering
\begin{dependency}[theme=simple, label style={font=\small}]
    \begin{deptext}[column sep=1ex]
    ROOT \& est \& un \& probléme  \\
    \& VRB \& DET \& NOM  \\
    \end{deptext}
    \depedge{1}{2}{root}
    \depedge{4}{3}{spe}
    \depedge[edge height=3ex]{2}{4}{dep}
\end{dependency}
\caption{
An example of a dependency tree.
Each word is annotated with its part-of-speech, head, and dependency relation.
}
\label{fig:dep-tree}
\end{figure}

\subsection{Dependency Parsing Module}

As illustrated in Figure~\ref{fig:overview}, CTC decoding results are used to obtain ``audio word embeddings'' and their dependency annotations.
Guided by the segmentation determined by CTC decoding, the corresponding segments of the speech representation matrix are treated as representations of individual words.
These representations are input to an LSTM to obtain audio word embeddings and then to Dep2label~\cite{strzyzViableDependencyParsing2019} for parsing.
Dep2label comprises a bi-LSTM with softmax, which computes the probability distribution of the three dependency annotations: part-of-speech (POS) tag, the relative position of the head, and the dependency relation.

\subsection{Handling ASR Errors in Training}
\label{ssec:oracle}

Since Wav2tree performs ASR before dependency parsing, the ASR output may contain errors, and predicting the correct parse tree becomes impossible.
%obtaining the gold parse tree in the presence of ASR errors becomes impossible.
Therefore, a corrective step is introduced during training to rewrite the parse tree according to the ASR error.
This provisional tree is referred to as \emph{an oracle}.
The oracle is obtained following two steps:

\begin{itemize}
    \item[(1)] Take an alignment between the gold transcription and the predicted words.
    \item[(2)] Rewrite a tree following the rules by Yoshikawa et al.~\cite{yoshikawaJointTransitionbasedDependency2016}
\end{itemize}

\begin{figure}[t]
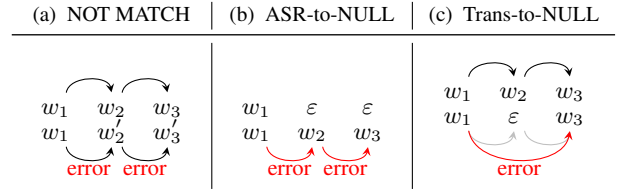

\centering
\begin{tabular}{c|c|c}
\footnotesize (a) \ NOT MATCH & \footnotesize (b) \ ASR-to-NULL & \footnotesize (c) \ Trans-to-NULL \\\midrule
\begin{dependency}[theme=simple, label style={font=\large}]
    \begin{deptext}[column sep=1.5ex]
    $w_1$ \& $w_2$ \& $w_3$  \\
    $w_1$ \& $w_2'$ \& $w_3'$  \\
    \end{deptext}
    \depedge{1}{2}{}
    \depedge{2}{3}{}
    \depedge[label style={below}, edge below]{1}{2}{\color{red} error}
    \depedge[label style={below}, edge below]{2}{3}{\color{red} error}
\end{dependency}
&
\begin{dependency}[theme=simple, label style={font=\large}]
    \begin{deptext}[column sep=1.5ex]
    $w_1$ \& $\varepsilon$ \& $\varepsilon$  \\
    $w_1$ \& $w_2$ \& $w_3$  \\
    \end{deptext}
    \depedge[label style={below}, edge below, color=red]{1}{2}{\color{red} error}
    \depedge[label style={below}, edge below, color=red]{2}{3}{\color{red} error}
\end{dependency}
&
\begin{dependency}[theme=simple, label style={font=\large}]
    \begin{deptext}[column sep=1.5ex]
    $w_1$ \& $w_2$ \& $w_3$  \\
    $w_1$ \& $\varepsilon$ \& $w_3$  \\
    \end{deptext}
    \depedge{1}{2}{}
    \depedge{2}{3}{}
    \depedge[label style={below}, edge below, color=lightgray]{1}{2}{}
    \depedge[label style={below}, edge below, color=lightgray]{2}{3}{}
    \depedge[label style={below}, edge below, color=red]{1}{3}{\color{red} error}
\end{dependency}
\end{tabular}
\caption{
The rules for obtaining the oracle tree.
The upper displays the gold dependency tree; the lower displays ASR results and oracles.
Annotations added after rewriting are highlighted in red; deleted are in gray.
}
\label{fig:rewrite}
\end{figure}

As described in Figure~\ref{fig:rewrite}, the rules in (2) involve the following three cases:

\begin{itemize}
\item[(a)] NOT MATCH: ASR error exists, but alignment is successful ($w_2\neq w_2'$, $w_3\neq w_3'$).
Rewrite the corresponding dependency relations to {\color{red} error}.
\item[(b)] ASR-to-NULL: There are excessive words in the ASR result.
Change the heads of those words to the previous word and attach {\color{red} error} relation.
\item[(c)] Trans-to-NULL: There are words missing from the ASR result.
Remove the edges attached to those words.
If there is a word whose head is a removed word ($w_3$ in Figure~\ref{fig:rewrite}), change its head to the head of the removed word, attaching {\color{red} error} relation.
\end{itemize}

\section{Textless Dependency Parsing}
\label{sec:ours}

This section describes our proposed method for textless dependency parsing.
The overview is shown on the right of Figure~\ref{fig:overview}.

The proposed method models $p(\mathbf{s}|\mathbf{X})$, where $\mathbf{s} = s_1 \ s_2 \ldots s_n$ is \emph{a labeled sequence} representing a dependency tree. 
This is achieved by directly predicting a labeled sequence, corresponding to a dependency parse tree, from speech representations without explicit ASR.
The prediction of a labeled sequence is learned straightforwardly using a CTC loss similar to the ASR module described in Section~\ref{ssec:asr}.
This method is inspired by previous work which showed improved ASR performance when jointly predicting linguistic annotations (phonemes and part-of-speech tags)~\cite{omachiEndtoendASRJointly2021}.
%the sequence-to-sequence model that improves ASR by predicting transcriptions and annotations (phonemes and part-of-speech tags) simultaneously~\cite{omachiEndtoendASRJointly2021}.

A labeled sequence is formed by concatenating words and their corresponding dependency annotations.
Figure~\ref{fig:seq} illustrates the labeled sequence of a dependency tree in Figure~\ref{fig:dep-tree}.
The annotations are mapped into special symbols enclosed in angle brackets: \texttt{<POS$j$>}, \texttt{<L$j$>} or \texttt{<R$j$>}, and \texttt{<REL$j$>}.
Hereafter, POS and dependency relations are written without mapping for ease of reading, such as \texttt{<VRB>} or \texttt{<dep>}.

\begin{figure*}[t]
\centering
\footnotesize
\begin{tabular}{ll}
\toprule
labeled sequence
&
\texttt{est<POS1><L1><REL0>\_un<POS2><R1><REL2>\_probléme<POS0><L2><REL1>}
\\\midrule
BPE-tokenized sequence
&
\texttt{est <POS1> <L1> <REL0> \_un <POS2> <R1> <REL2> \_prob lé me <POS0> <L2> <REL1>}
\\
\bottomrule
\end{tabular}
\caption{A labeled sequence representing a dependency tree in Figure~\ref{fig:dep-tree} and its BPE tokenization.
Spaces are indicated by ``\_''.}
\label{fig:seq}
\end{figure*}

\subsection{Recovering Dependency Tree from Labeled Sequence}

Given the predicted labeled sequence $s_1 \ s_2 \ \ldots \ s_n$ (where each $s_i$ is obtained by splitting with space symbols), it is required to define the way to recover dependency tree annotations from it.
For each $s_i$, dependency annotations $w_i, p_i, h_i, r_i$ (each represents word, POS, relative position of the head, and dependency relation, respectively) are determined by the following rule:

\begin{enumerate}
\item $w_i$ is a sequence up to just before the leftmost symbol ``\texttt{<}''.
\item $p_i$, $h_i$ and $r_i$ are mapped from the leftmost labels. For example, $p_i$ is mapped from the leftmost ``\texttt{<POS$j$>}''.
\item If the annotation is not assigned in 2 (no labels were found), assign generic labels: $p_i=\mathrm{X}$, $h_i=\mathrm{None}$, $r_i=\mathrm{dep}$.
\end{enumerate}
Here, $h_i=\mathrm{None}$ indicates that $w_i$ does not have a head.

Similar to \cite{pupierEndtoEndDependencyParsing2022}, we impose three constraints on the dependency structure: (1) uniqueness of root, (2) uniqueness of head, and (3) acyclicity.
Hence, we perform heuristic post-processing  proposed in \cite{strzyzViableDependencyParsing2019} to guarantee these constraints.

% \begin{itemize}
% \item root が存在しない場合，先頭の単語を root とする (1)
% \item root が複数存在する場合，そのうち一番左の単語 $w_L$ を root とし，それ以外の単語の主辞を $w_L$ に変更する (1)
% \item 主辞のない単語の主辞を root の単語とする (2)
% \item 巡回の一番左の主辞を root の単語とする (3)
% \end{itemize}

\section{Experimental Settings}

\subsection{Dataset}

We evaluate each model on two languages: French and English.
The French dataset is obtained from Orféo Treebank~\cite{benzitounProjetORFEOCorpus2016}, which is also used in Wav2tree~\cite{pupierEndtoEndDependencyParsing2022}.
Orféo Treebank is a collection of the corpus with both speech audio and corresponding dependency tree annotations.
For English, the dataset is obtained from the Switchboard Telephone Speech Corpus~\cite{godfreySWITCHBOARDTelephoneSpeech1992}.
Since the Switchboard corpus includes gold phrase structure annotations, we converted them into dependency trees (described in Section \ref{ssec:prep}).
Note that the dependency tree annotations from Orféo Treebank may contain errors, as a significant portion (95 \%) of them are generated using an off-the-shelf parser~\cite{nasrAnnotationSyntaxiqueAutomatique2020}.

\subsubsection{Preprocessing}
\label{ssec:prep}

For Orféo Treebank, we used the dataset available in Wav2tree repository\footnote{\url{https://gricad-gitlab.univ-grenoble-alpes.fr/pupiera/Wav2tree_release}}, so no additional preprocessing is required.
To create a dataset from the Switchboard corpus, we extracted phrase structures and time ranges of each sentence from NXT Switchboard Annotations~\cite{calhounNXTformatSwitchboardCorpus2010}.
Similar to~\cite{honnibalJointIncrementalDisfluency2014}, we converted phrase structures to dependency trees using Stanford dependency converter~\cite{demarneffeGeneratingTypedDependency2006}, and nodes referring to punctuation and meta information (e.g. end-of-sentence) are removed.
Table~\ref{tab:dataset} shows the statistics of the dataset we finally obtained.

\begin{table}[t]
\centering
\caption{Statistics of the dataset}
\label{tab:dataset}
\footnotesize
\begin{tabular}{l|l|lll}
\toprule
Corpus & Statistic & Train & Dev & Test \\\midrule
\multirow{3}{*}{
\begin{tabular}{l}
Orféo Treebank\\
(French, \cite{benzitounProjetORFEOCorpus2016})
\end{tabular}
}
& Size & 169,505 & 21,301 & 21,459
\\
& Duration & 130.9h & 16.6h & 16.8h
\\
& Avg. Words & 10.1 & 10.2 & 10.2
\\\midrule
\multirow{3}{*}{
\begin{tabular}{l}
Switchboard\\
(English, \cite{godfreySWITCHBOARDTelephoneSpeech1992})
\end{tabular}
}
& Size & 61964 & 7810 & 7771 \\
& Duration & 48.5h & 6.1h & 6.3h \\
& Avg. Words & 10.2 & 10.2 & 10.4
\\\bottomrule
\end{tabular}
\end{table}

\subsection{Model}

As a feature extractor $f$, we used a \texttt{LeBenchmark/wav2}  
\texttt{vec2-FR-7K-large}~\cite{evainLeBenchmarkReproducibleFramework2021} for the French model and \texttt{face}  
\texttt{book/wav2vec2-large-robust}~\cite{hsuRobustWav2vecAnalyzing2021} for the English model, and updated their parameters during training.
The FNN for obtaining the speech representation comprises three fully connected layers of the dimension size $d=1024$, a dropout ratio of $0.15$, and employs layer normalization and Leaky ReLu activation functions.
We used Adadelta optimizer~\cite{zeilerADADELTAAdaptiveLearning2012} with a learning rate of 1.0.
The models are trained for 30 epochs.

As $\textrm{Dec}_{\textrm{ctc}}$, we employed a CTC greedy decoder.
For $\textrm{Dec}_{\textrm{spm}}$, we generated SentencePiece vocabulary $\mathcal{V}$ using Byte-Pair Encoding (BPE, \cite{kudoSentencePieceSimpleLanguage2018}) of vocabulary size 1000.
In the textless method, special label tokens (such as \texttt{POS$j$}) are added to the vocabulary as \texttt{user\_defined\_symbols}.

\subsection{Optimization}

Models were trained on a single NVIDIA A100.
The proposed model has fewer parameters than the existing model due to the lack of a dedicated network for parsing.  Additionally, the training time for the proposed method is shorter compared to Wav2tree, as our method does not involve rewriting the gold dependency tree, which consumes a significant portion of the training time.

\subsection{Evaluation}

We evaluated the performance of models from two aspects: ASR metrics and parsing metrics.
The former includes WER and CER.
The latter includes POS accuracy, UAS (unlabeled attachment score), and LAS (labeled attachment score).
In evaluating parsing performance, we rewrite the predicted tree following the rules described in Section~\ref{ssec:oracle}.

\begin{table*}[t]
\centering
\caption{Experimental Result. Training time is the average of the first 10 epochs.}
\label{tab:result}
\begin{tabular}{l|l||cc|ccc||cc}
\toprule
  Corpus & Model & \multicolumn{2}{c|}{\footnotesize ASR Metrics $\downarrow$} & \multicolumn{3}{c||}{\footnotesize Parsing Metrics $\uparrow$} & \multicolumn{2}{c}{\footnotesize Model Comparison} \\
         & & WER  & CER  & POS & UAS  & LAS  & Parameters & Training time \\\midrule
\multirow{2}{*}{
\begin{tabular}{l}
Orféo Treebank (French)
\end{tabular}
}
& Textless  & 28.4 & 19.3 & 77.2 & 68.6 & 64.5 & 320M & 1:18 \\ % DONE: layer_lr2-1.0
& Wav2tree & 26.0 & 18.1 & 78.4 & 72.6 & 68.7 & 350M & 2:48 \\ % DONE: layer_lr2-1.0
\midrule
\multirow{2}{*}{
\begin{tabular}{l}
Switchboard (English)
\end{tabular}
}
& Textless & 10.3 & 5.6 & 90.9 & 79.7 & 75.7 & 320M & 0:26 \\ % DONE: layer_lr2-1.0
& Wav2tree & 9.7 & 5.2 & 91.3 & 84.1 & 79.8 & 353M & 1:05 \\ % DONE: layer_lr2-1.0
\bottomrule
\end{tabular}
\end{table*}

\section{Result and Discussion}

Table~\ref{tab:result} shows the experimental result.
Overall, Wav2tree outperforms the textless method both in ASR and parsing metrics.
Note that the English result is significantly better than the French result, even though the dataset size is nearly three times smaller.
This discrepancy could be attributed to the fact that the model (\texttt{facebook/wav2vec2-large-robust}) was pretrained on Switchboard as well, which enhances ASR performance and leads to improved parsing accuracy.

\subsection{Analysis 1: Prediction Accuracy of Head Position}

As a reason for the superior parsing accuracy of Wav2tree, we hypothesize that the resolution of the longer-distance dependencies requires word-level representations.
To test this hypothesis, we calculated the prediction accuracy of the head position for four representative POS tags (ADJ, ADV, NOUN, and VERB\footnote{In Orféo Treebank, NOUN and VERB are annotated as NOM and VRB, respectively.}).
Alongside the accuracy metrics, we report the co-occurrence frequencies of POS tags and the head positions for reference.

\begin{figure}[t]
\centering
\includegraphics[width=.47\textwidth]{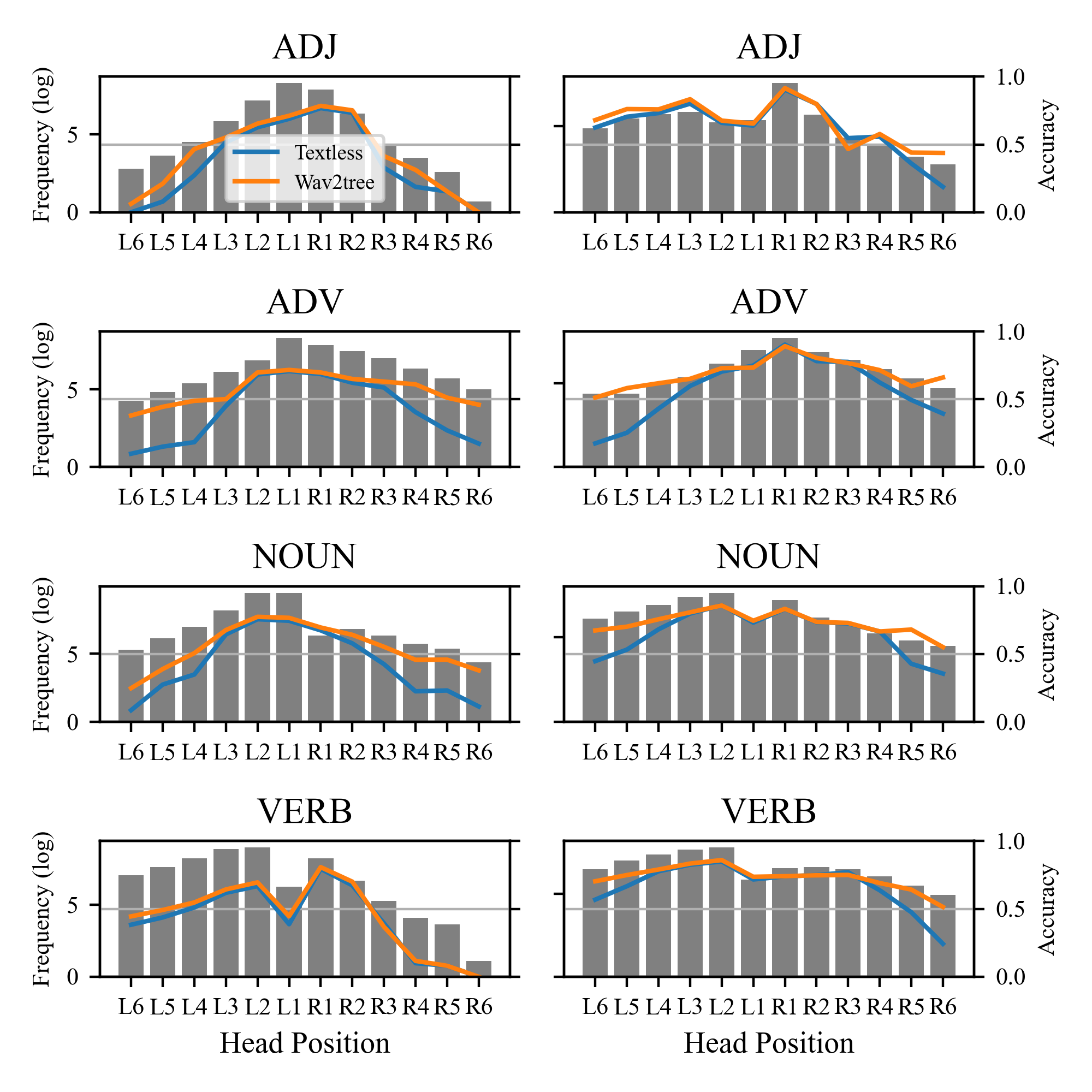}
\caption{
Prediction accuracy of the relative position of the head (left: Orféo Treebank, right: Switchboard).
Bars show log frequencies; lines show accuracies.
}
\label{fig:pos_acc}
\end{figure}

Figure~\ref{fig:pos_acc} shows the result.
This observation supports the hypothesis that explicitly segmenting a speech at the word boundary is crucial in predicting long-distance dependencies.

\begin{table}[t]
\centering
\caption{
Instances where the textless method predicted correctly, while Wav2tree did not. 
Words emphasized with stress (higher intensity or pitch) are highlighted in bold.
}
\label{tab:tree}
\begin{tabular}{cc}
\toprule
\footnotesize{Gold / Prediction (Textless) \ {\color{green}\cmark}}
&
\footnotesize{Prediction (Wav2tree) \ {\color{red}\xmark}}
\\\midrule
\begin{dependency}[theme=simple, label style={font=\small}]
    \scriptsize
    \begin{deptext}[column sep=0.3ex]
    go \& \textbf{buy} \& me \& some \& strawberries \\
    \end{deptext}
    \deproot[edge unit distance=1.5ex]{2}{}
    \depedge{2}{1}{}
    \depedge{2}{3}{}
    \depedge{5}{4}{}
    \depedge{2}{5}{}
\end{dependency}
&
\begin{dependency}[theme=simple, label style={font=\small}]
    \scriptsize
    \begin{deptext}[column sep=0.3ex]
    go \& \textbf{buy} \& me \& some \& strawberries \\
    \end{deptext}
    \deproot[edge unit distance=1.5ex]{1}{}
    \depedge{1}{2}{}
    \depedge{2}{3}{}
    \depedge{5}{4}{}
    \depedge{2}{5}{}
\end{dependency}
\\\midrule
\begin{dependency}[theme=simple, label style={font=\small}]
    \scriptsize
    \begin{deptext}[column sep=0.4ex]
    but \& it \& 's \& just \& wide \& \textbf{open} \\
    \end{deptext}
    \deproot[edge unit distance=1.5ex]{6}{}
    \depedge{6}{1}{}
    \depedge{6}{2}{}
    \depedge{6}{3}{}
    \depedge{6}{4}{}
    \depedge{6}{5}{}
\end{dependency}
&
\begin{dependency}[theme=simple, label style={font=\small}]
    \scriptsize
    \begin{deptext}[column sep=0.4ex]
    but \& it \& 's \& just \& wide \& \textbf{open} \\
    \end{deptext}
    \deproot[edge unit distance=1.5ex]{6}{}
    \depedge{5}{1}{}
    \depedge{5}{2}{}
    \depedge{5}{3}{}
    \depedge{5}{4}{}
    \depedge{6}{5}{}
\end{dependency}
\\\midrule
\begin{dependency}[theme=simple, label style={font=\small}]
    \scriptsize
    \begin{deptext}[column sep=0.4ex]
    i \& went \& \textbf{horseback} \& riding \\
    \end{deptext}
    \deproot[edge unit distance=1.5ex]{2}{}
    \depedge{2}{1}{}
    \depedge{4}{3}{}
    \depedge{2}{4}{}
\end{dependency}
&
\begin{dependency}[theme=simple, label style={font=\small}]
    \scriptsize
    \begin{deptext}[column sep=0.4ex]
    i \& went \& \textbf{horseback} \& riding \\
    \end{deptext}
    \deproot[edge unit distance=1.5ex]{2}{}
    \depedge{2}{1}{}
    \depedge{2}{3}{}
    \depedge{2}{4}{}
\end{dependency}
\\
\bottomrule
\end{tabular}
\end{table}

\subsection{Analysis 2: Advantage of Textless Method}

To discern the relative advantages of the textless method, we investigate the cases where the textless method predicts better than Wav2tree.
To this end, we collected instances where UAS of the textless method is 1.0 and that of Wav2tree is below the average, and without word prediction errors.
We obtained 40 instances in total from Switchboard dataset.

Among them, we found six instances where the stressed pronunciation appears to aid in accurate parsing.
Table~\ref{tab:tree} shows concrete examples.
In the first example, with two candidates for the root word (``go'' and ``buy''), the textless method accurately predicts the correct one (``buy'') which exhibits higher intensity and pitch.
In the second example, while the correct complement of ``it'' is ``open'', it is also possible to mistakenly recognize it as ``wide'', considering the partial phrase ``it's just wide''.
Here Wav2tree makes the wrong prediction, while the textless method correctly identifies the stressed ``open'' as the complement.
In the third example, the stressed pronunciation of ``horseback'' elucidates that ``horseback riding'' is a compound word. 
This means that the head of ``horseback'' is ``riding'', not ``went''.
This structure is correctly predicted by the textless method and not by Wav2tree.

Each of these instances exemplifies cases where the sentence-level prosodic contour is crucial for making the correct prediction.
We conjecture that the proposed method was able to successfully parse these utterances by modeling the prosody of the whole sentence.
In contrast, since Wav2tree performs parsing with word representations embedded independently, it may fail to capture a prosodic contour of the whole sentence.
This suggests the necessity for leveraging sentence-level prosody to enhance parsing performance further on spoken audio.

% Limitations
While this analysis presents a fragment of evidence supporting the positive effect of the sentence-level prosody for parsing, which is in line with previous arguments \cite{grosjeanPatternsSilencePerformance1979, priceUseProsodySyntactic1991}, the causal relationship between sentence-level prosody and syntactic disambiguation remains unclear.
For future work, it is beneficial to construct an evaluation set targeting syntactic disambiguation with the audio feature.

\section{Conclusion}

In this work, we proposed a method for textless dependency parsing from a speech signal and examined its effectiveness and limitations.
Through the comparative experiment, we suggest the contribution of word-level representations, particularly in predicting long-distance dependency relationships.
Besides, we found that the proposed textless method works well when the distinct audio features (such as higher intensity or pitch) seem to help parsing, suggesting the contribution of the sentence-level prosody in parsing.
Our findings highlight the importance of integrating both word-level representations and sentence-level prosody to enhance parsing performance further in speech.
Our method has a limitation in that it is based solely on CTC, which assumes conditional independence.
Future work could explore the effect of attention mechanisms~\cite{kimJointCTCattentionBased2017} or intermediate CTC architectures~\cite{nozakiRelaxingConditionalIndependence2021} to overcome such limitations.

\section{Acknowledgements}
This work was supported by JST Moonshot JPMJMS2237 and JST FOREST JPMJFR226V.

\bibliographystyle{IEEEtran}
\bibliography{interspeech2024-submission}

% Generated by IEEEtran.bst, version: 1.13 (2008/09/30)
\begin{thebibliography}{10}
\providecommand{\url}[1]{#1}
\csname url@samestyle\endcsname
\providecommand{\newblock}{\relax}
\providecommand{\bibinfo}[2]{#2}
\providecommand{\BIBentrySTDinterwordspacing}{\spaceskip=0pt\relax}
\providecommand{\BIBentryALTinterwordstretchfactor}{4}
\providecommand{\BIBentryALTinterwordspacing}{\spaceskip=\fontdimen2\font plus
\BIBentryALTinterwordstretchfactor\fontdimen3\font minus \fontdimen4\font\relax}
\providecommand{\BIBforeignlanguage}[2]{{%
\expandafter\ifx\csname l@#1\endcsname\relax
\typeout{** WARNING: IEEEtran.bst: No hyphenation pattern has been}%
\typeout{** loaded for the language `#1'. Using the pattern for}%
\typeout{** the default language instead.}%
\else
\language=\csname l@#1\endcsname
\fi
#2}}
\providecommand{\BIBdecl}{\relax}
\BIBdecl

\bibitem{lakhotiaGenerativeSpokenLanguage2021}
\BIBentryALTinterwordspacing
K.~Lakhotia, E.~Kharitonov, W.-N. Hsu, Y.~Adi, A.~Polyak, B.~Bolte, T.-A. Nguyen, J.~Copet, A.~Baevski, A.~Mohamed, and E.~Dupoux, ``On {{Generative Spoken Language Modeling}} from {{Raw Audio}},'' \emph{Transactions of the Association for Computational Linguistics}, vol.~9, pp. 1336--1354, 2021.
\BIBentrySTDinterwordspacing

\bibitem{polyakSpeechResynthesisDiscrete2021}
\BIBentryALTinterwordspacing
A.~Polyak, Y.~Adi, J.~Copet, E.~Kharitonov, K.~Lakhotia, W.-N. Hsu, A.~Mohamed, and E.~Dupoux, ``Speech {{Resynthesis}} from {{Discrete Disentangled Self-Supervised Representations}},'' in \emph{Interspeech 2021}.\hskip 1em plus 0.5em minus 0.4em\relax {ISCA}, 2021, pp. 3615--3619.
\BIBentrySTDinterwordspacing

\bibitem{kreukTextlessSpeechEmotion2022}
\BIBentryALTinterwordspacing
F.~Kreuk, A.~Polyak, J.~Copet, E.~Kharitonov, T.~A. Nguyen, M.~Rivière, W.-N. Hsu, A.~Mohamed, E.~Dupoux, and Y.~Adi, ``Textless {{Speech Emotion Conversion}} using {{Discrete}} \& {{Decomposed Representations}},'' in \emph{Proceedings of the 2022 {{Conference}} on {{Empirical Methods}} in {{Natural Language Processing}}}, Y.~Goldberg, Z.~Kozareva, and Y.~Zhang, Eds.\hskip 1em plus 0.5em minus 0.4em\relax {Association for Computational Linguistics}, 2022, pp. 11\,200--11\,214.
\BIBentrySTDinterwordspacing

\bibitem{pupierEndtoEndDependencyParsing2022}
\BIBentryALTinterwordspacing
A.~Pupier, M.~Coavoux, B.~Lecouteux, and J.~Goulian, ``End-to-{{End Dependency Parsing}} of {{Spoken French}},'' in \emph{Interspeech 2022}.\hskip 1em plus 0.5em minus 0.4em\relax {ISCA}, 2022, pp. 1816--1820.
\BIBentrySTDinterwordspacing

\bibitem{omachiEndtoendASRJointly2021}
\BIBentryALTinterwordspacing
M.~Omachi, Y.~Fujita, S.~Watanabe, and M.~Wiesner, ``End-to-end {{ASR}} to jointly predict transcriptions and linguistic annotations,'' in \emph{Proceedings of the 2021 {{Conference}} of the {{North American Chapter}} of the {{Association}} for {{Computational Linguistics}}: {{Human Language Technologies}}}.\hskip 1em plus 0.5em minus 0.4em\relax {Association for Computational Linguistics}, 2021, pp. 1861--1871.
\BIBentrySTDinterwordspacing

\bibitem{baevskiWav2vecFrameworkSelfSupervised2020}
\BIBentryALTinterwordspacing
A.~Baevski, Y.~Zhou, A.~Mohamed, and M.~Auli, ``Wav2vec 2.0: {{A Framework}} for {{Self-Supervised Learning}} of {{Speech Representations}},'' in \emph{Advances in {{Neural Information Processing Systems}}}, vol.~33.\hskip 1em plus 0.5em minus 0.4em\relax {Curran Associates, Inc.}, 2020, pp. 12\,449--12\,460.
\BIBentrySTDinterwordspacing

\bibitem{gravesConnectionistTemporalClassification}
\BIBentryALTinterwordspacing
A.~Graves, S.~Fern\'{a}ndez, F.~Gomez, and J.~Schmidhuber, ``Connectionist temporal classification: labelling unsegmented sequence data with recurrent neural networks,'' in \emph{Proceedings of the 23rd International Conference on Machine Learning}, ser. ICML '06.\hskip 1em plus 0.5em minus 0.4em\relax New York, NY, USA: Association for Computing Machinery, 2006, p. 369–376.
\BIBentrySTDinterwordspacing

\bibitem{kudoSentencePieceSimpleLanguage2018}
\BIBentryALTinterwordspacing
T.~Kudo and J.~Richardson, ``{{SentencePiece}}: {{A}} simple and language independent subword tokenizer and detokenizer for {{Neural Text Processing}},'' in \emph{Proceedings of the 2018 {{Conference}} on {{Empirical Methods}} in {{Natural Language Processing}}: {{System Demonstrations}}}, E.~Blanco and W.~Lu, Eds.\hskip 1em plus 0.5em minus 0.4em\relax {Association for Computational Linguistics}, 2018, pp. 66--71.
\BIBentrySTDinterwordspacing

\bibitem{strzyzViableDependencyParsing2019}
\BIBentryALTinterwordspacing
M.~Strzyz, D.~Vilares, and C.~Gómez-Rodríguez, ``Viable {{Dependency Parsing}} as {{Sequence Labeling}},'' in \emph{Proceedings of the 2019 {{Conference}} of the {{North American Chapter}} of the {{Association}} for {{Computational Linguistics}}: {{Human Language Technologies}}, {{Volume}} 1 ({{Long}} and {{Short Papers}})}.\hskip 1em plus 0.5em minus 0.4em\relax {Association for Computational Linguistics}, 2019, pp. 717--723.
\BIBentrySTDinterwordspacing

\bibitem{yoshikawaJointTransitionbasedDependency2016}
\BIBentryALTinterwordspacing
M.~Yoshikawa, H.~Shindo, and Y.~Matsumoto, ``Joint {{Transition-based Dependency Parsing}} and {{Disfluency Detection}} for {{Automatic Speech Recognition Texts}},'' in \emph{Proceedings of the 2016 {{Conference}} on {{Empirical Methods}} in {{Natural Language Processing}}}, J.~Su, K.~Duh, and X.~Carreras, Eds.\hskip 1em plus 0.5em minus 0.4em\relax {Association for Computational Linguistics}, 2016, pp. 1036--1041.
\BIBentrySTDinterwordspacing

\bibitem{benzitounProjetORFEOCorpus2016}
\BIBentryALTinterwordspacing
C.~Benzitoun, J.-M. Debaisieux, and H.-J. Deulofeu, ``Le projet orfÉo : un corpus d’étude pour le français contemporain,'' no.~15, 2016.
\BIBentrySTDinterwordspacing

\bibitem{godfreySWITCHBOARDTelephoneSpeech1992}
\BIBentryALTinterwordspacing
J.~Godfrey, E.~Holliman, and J.~McDaniel, ``{{SWITCHBOARD}}: Telephone speech corpus for research and development,'' in \emph{{{IEEE International Conference}} on {{Acoustics}}, {{Speech}} and {{Signal Processing}} ({{ICASSP}})}, vol.~1, 1992, pp. 517--520 vol.1.
\BIBentrySTDinterwordspacing

\bibitem{nasrAnnotationSyntaxiqueAutomatique2020}
\BIBentryALTinterwordspacing
A.~Nasr, F.~Dary, F.~Béchet, and B.~Fabre, ``Annotation syntaxique automatique de la partie orale du orfÉo,'' vol. 219, no.~3, pp. 87--102, 2020.
\BIBentrySTDinterwordspacing

\bibitem{calhounNXTformatSwitchboardCorpus2010}
\BIBentryALTinterwordspacing
S.~Calhoun, J.~Carletta, J.~M. Brenier, N.~Mayo, D.~Jurafsky, M.~Steedman, and D.~Beaver, ``The {{NXT-format Switchboard Corpus}}: A rich resource for investigating the syntax, semantics, pragmatics and prosody of dialogue,'' vol.~44, no.~4, pp. 387--419, 2010.
\BIBentrySTDinterwordspacing

\bibitem{honnibalJointIncrementalDisfluency2014}
\BIBentryALTinterwordspacing
M.~Honnibal and M.~Johnson, ``Joint {{Incremental Disfluency Detection}} and {{Dependency Parsing}},'' \emph{Transactions of the Association for Computational Linguistics}, vol.~2, pp. 131--142, 2014.
\BIBentrySTDinterwordspacing

\bibitem{demarneffeGeneratingTypedDependency2006}
\BIBentryALTinterwordspacing
M.-C. de~Marneffe, B.~MacCartney, and C.~D. Manning, ``Generating {{Typed Dependency Parses}} from {{Phrase Structure Parses}},'' in \emph{Proceedings of the {{Fifth International Conference}} on {{Language Resources}} and {{Evaluation}} ({{LREC}}'06)}, N.~Calzolari, K.~Choukri, A.~Gangemi, B.~Maegaard, J.~Mariani, J.~Odijk, and D.~Tapias, Eds.\hskip 1em plus 0.5em minus 0.4em\relax {European Language Resources Association (ELRA)}, 2006.
\BIBentrySTDinterwordspacing

\bibitem{evainLeBenchmarkReproducibleFramework2021}
\BIBentryALTinterwordspacing
S.~Evain, H.~Nguyen, H.~Le, M.~Z. Boito, S.~Mdhaffar, S.~Alisamir, Z.~Tong, N.~Tomashenko, M.~Dinarelli, T.~Parcollet, A.~Allauzen, Y.~Esteve, B.~Lecouteux, F.~Portet, S.~Rossato, F.~Ringeval, D.~Schwab, and L.~Besacier, ``{{LeBenchmark}}: {{A Reproducible Framework}} for {{Assessing Self-Supervised Representation Learning}} from {{Speech}},'' in \emph{Interspeech 2021}, 2021, pp. 1439--1443.
\BIBentrySTDinterwordspacing

\bibitem{hsuRobustWav2vecAnalyzing2021}
\BIBentryALTinterwordspacing
W.-N. Hsu, A.~Sriram, A.~Baevski, T.~Likhomanenko, Q.~Xu, V.~Pratap, J.~Kahn, A.~Lee, R.~Collobert, G.~Synnaeve, and M.~Auli, ``Robust wav2vec 2.0: {{Analyzing Domain Shift}} in {{Self-Supervised Pre-Training}},'' 2021, pp. 721--725.
\BIBentrySTDinterwordspacing

\bibitem{zeilerADADELTAAdaptiveLearning2012}
\BIBentryALTinterwordspacing
M.~D. Zeiler. (2012) {{ADADELTA}}: {{An Adaptive Learning Rate Method}}.
\BIBentrySTDinterwordspacing

\bibitem{grosjeanPatternsSilencePerformance1979}
\BIBentryALTinterwordspacing
F.~Grosjean, L.~Grosjean, and H.~Lane, ``The patterns of silence: {{Performance}} structures in sentence production,'' vol.~11, no.~1, pp. 58--81, 1979.
\BIBentrySTDinterwordspacing

\bibitem{priceUseProsodySyntactic1991}
\BIBentryALTinterwordspacing
P.~Price, M.~Ostendorf, S.~Shattuck-Hufnagel, and C.~Fong, ``The {{Use}} of {{Prosody}} in {{Syntactic Disambiguation}},'' in \emph{Speech and {{Natural Language}}: {{Proceedings}} of a {{Workshop Held}} at {{Pacific Grove}}, {{California}}, {{February}} 19-22}, 1991.
\BIBentrySTDinterwordspacing

\bibitem{kimJointCTCattentionBased2017}
\BIBentryALTinterwordspacing
S.~Kim, T.~Hori, and S.~Watanabe, ``Joint {{CTC-attention}} based end-to-end speech recognition using multi-task learning,'' in \emph{{{IEEE International Conference}} on {{Acoustics}}, {{Speech}} and {{Signal Processing}} ({{ICASSP}})}, 2017, pp. 4835--4839.
\BIBentrySTDinterwordspacing

\bibitem{nozakiRelaxingConditionalIndependence2021}
\BIBentryALTinterwordspacing
J.~Nozaki and T.~Komatsu, ``Relaxing the {{Conditional Independence Assumption}} of {{CTC-Based ASR}} by {{Conditioning}} on {{Intermediate Predictions}},'' in \emph{Interspeech 2021}.\hskip 1em plus 0.5em minus 0.4em\relax {ISCA}, 2021, pp. 3735--3739.
\BIBentrySTDinterwordspacing

\end{thebibliography}

\end{document}